\documentclass[11pt]{article}
\usepackage{amsmath}
\usepackage{amsfonts}
\usepackage{amssymb} 
\usepackage{amsthm}
\usepackage{mathtools}
\usepackage{mathabx}
\usepackage{authblk}
\usepackage{xspace} 
\usepackage{tabu}
\usepackage[export]{adjustbox}
\usepackage{graphicx} 
\usepackage[lined]{algorithm2e}
\usepackage{tikz}
\usepackage{tkz-graph}
\usetikzlibrary{arrows, shapes, patterns}
\newif\ifcomment
\commenttrue

\newcommand{\tuple}[1]{\langle #1 \rangle}
\newcommand{\Bool}{{\mathbb B}}
\newcommand{\Bset}{{\mathbb B}}

\newcommand{\abbrev}[1]{#1, \relax}
\newcommand{\ie}[0]{\abbrev{\textit{ie.}}}

\newcommand{\taboon}{\textsc{t}a\textsc{b}oo\textsc{n}\xspace}

\newcommand{\evo}[0]{\longrightarrow}
\newcommand{\xevo}[1]{\stackrel{#1}{\evo}}
\newcommand{\arc}{\ensuremath{\, \tikz[baseline=-0.5ex, scale=0.65, thin]{\draw[->, >=open triangle 45](0, 0)to[bend left=0] (1, 0);}\, }}
\newcommand{\blackarc}{\ensuremath{\, \tikz[baseline=-0.5ex, scale=0.65, thin]{\draw[->, >= triangle 45](0, 0)to[bend left=0] (1, 0);}\, }}
\newcommand{\sarc}[1]{\stackrel{\kern-1ex{#1}}{\arc}}
\newcommand{\sblackarc}[1]{\stackrel{\kern-1ex{#1}}{\blackarc}}

\newcommand{\eqdef}{\overset{\textrm{def}}{=}}

\newcommand{\signed}[1]{\ensuremath{\, \tikz[baseline=-0.5ex, scale=0.65, thin]{\draw[->, >=open triangle 45](0, 0)to[bend left=0] (1, 0); \node at (0.4, 0.3){#1};
}\, }}

\begin{document}
\title{
	TaBooN \\
	Boolean Network Synthesis Based on Tabu Search}
\author[1]{Sara Sadat Aghamiri}
 \author[2]{Franck Delaplace \thanks{Corresponding author}}

\affil[1,2]{Paris-Saclay University -Univ. Evry, IBISC laboratory}
\date{}

\maketitle
\begin{abstract}
 Recent developments in Omics-technologies revolutionized the investigation of biology by producing molecular data in multiple dimensions and scale. This breakthrough in biology raises the crucial issue of their interpretation based on modelling. In this undertaking, network provides a suitable framework for modelling the interactions between molecules. Basically a Biological network is composed of nodes referring to the components such as genes or proteins, and the edges/arcs formalizing interactions between them. The evolution of the interactions is then modelled by the definition of a dynamical system. Among the different categories of network, the Boolean network offers a reliable qualitative framework for the modelling. Automatically synthesizing a Boolean network from experimental data therefore remains a necessary but challenging issue.

In this study, we present \taboon, an original work-flow for synthesizing Boolean Networks from biological data. The methodology uses the data in the form of Boolean profiles for inferring all the potential local formula inference. They combine to form the model space from which the most truthful model with regards to biological knowledge and experiments must be found. 
In the \taboon work-flow the selection of the fittest model is achieved by a Tabu-search algorithm.
\taboon is an automated method for Boolean Network inference from experimental data that can also assist to evaluate and optimize the dynamic behaviour of the biological networks providing a reliable platform for further modelling and predictions.
\end{abstract}

\section{Introduction}

Boolean networks (BNs) study applied to biology has been pioneered by S.~Kauffman \cite{glass1973logical, kauffman1969metabolic} and R.~Thomas \cite{thieffry1995dynamical, thomas1995dynamical} as a regulation network modelling approach. The BNs consist of a logical dynamical system formalizing the interactions between the elements of a regulatory network in terms of discrete variables, logical functions and parameters. Biological studies benefit from such networks to structure and represent the molecular interactions with the synthesis of BNs where the nodes refer to the components such as genes or proteins, and the edges show the interactions between them \cite{barabasi2011network, saadatpour2013Boolean, albert2008Boolean}. Also different expressions and productions of the molecules in various conditions can be characterized as on/active/up-regulated or off/inactive/down-regulated in BNs. BNs have been considered as a reliable standard approach to study signaling and regulatory networks, by modelling and analysing biological processes~\cite{faure2006dynamical, fumia2013Boolean, schlatter2009off, davidich2008Boolean, gupta2007Boolean}.

In the recent decades, technological advancements in data generation have produced unprecedented accelerating amounts of experimental data~\cite{patel2012ngs}. These technologies revolutionized the investigation of biology and human health, producing data in multiple dimensions and scale including DNA sequence, epigenomic states, single-cell gene expression activity, proteomics, functional, and phenotypic measurements~\cite{mardis2008next, wang2009rna}. The biological information accumulated over the years in databases and literature as a static source of knowledge \cite{mazein2018systems, dorier2016Boolean} that however provides limited insight into the system’s response to perturbations \cite{khatri2012ten}.

Although the availability of biological information plays a crucial role in the construction of biological networks, the procedure of synthesizing a biological BN manually can take a considerable amount of time and could be error prone. Indeed, in a manual construction of a regulatory network requires an iterated improvement of the network, validated by experimental data and biological knowledge that could reveal daunted due to the successive revisions for fitting  with the validation elements. Therefore, automating this procedure based on experimental data for synthesizing a biological BN, is an essential approach~\cite{lee2009computational}. 

Different approaches have been applied to automate and optimize the construction of these biological BNs.
In \cite{terfve2012cellnoptr} researchers proposed a pipeline for implementing a Boolean logic model from a ”prior knowledge network” (PKN, \ie, a network obtained from literature or expert knowledge) and trains it upon perturbation data. The \cite{terfve2012cellnoptr} analysis includes the import of the network and data, processing the network, training, and reporting the analysis results. This method can features different logic formalisms from Boolean models to differential equations in a common framework.

The approach to generate and optimize Boolean networks, based on a given PKN is also used by \cite{dorier2016Boolean}. This method utilizes an optimization approach to produce specific, contextualized models from generic PKNs. This procedure includes implementing a genetic algorithm to construct a model network as a sub-network of the PKN and trained against experimental data to reproduce the experimentally observed behaviour in terms of attractors and the transitions due to specific perturbations. The resulting model network forms a dynamic Boolean model that is more similar to the observed biological process used to train the model than the original PKN.

Other methodologies in this field \cite{chevalier2019synthesis} addresses the construction of Boolean functions from constraints on their domain and emerging dynamic properties of the resulting network. The existence and absence of trajectories between partially observed configurations, and stable behaviours (fixpoints and cyclic attractors) are associated with the dynamic properties. The construction of BNs in \cite{chevalier2019synthesis} expressed as a Boolean satisfiability problem relying on Answer-Set Programming with a parametrized complexity leading to a complete non-redundant characterization of the set of solutions.

Applying Answer Set Programming is also used by \cite{ostrowski2016Boolean}. Describing that an efficient and scalable training method focuses on the comparison of two time-points and assumes that the system has reached an early steady-state, \cite{ostrowski2016Boolean} generalizes such a learning procedure to take into account the time series traces of phosphoproteomics data to discriminate Boolean networks according to their transient dynamics. This identifies a specific condition that must be satisfied by the dynamics of a Boolean network to be compatible with a discretized time series trace. The methodology includes an Answer Set Programming to compute an over-approximation of the Boolean network set that fits best with experimental data and provides the corresponding encodings. This procedure, combined with model-checking approaches, points to a global learning algorithm. \\
In this field, another approach described by \cite{yordanov2016method} utilizes Satisfiability Modulo Theory. This method defines automated formal reasoning, which permits the construction and analysis of the complete set of logical models consistent with experimental observations. The procedure of this methodology includes identifying critical network components, defining definite and possible interactions, characterizing an Abstract Boolean Network, encoding experimental observations as constraints on state trajectories, enumerating the concrete models that satisfy these constraints, identifying minimal networks with specific features, along with prediction steps. As a result, this methodology transforms knowledge of complex biological processes from sets of possible interactions and experimental observations to predictive biological programs governing cell function.

In \cite{barman2018Boolean}, the authors defines a Boolean network inference method from time-series gene expression data using a genetic algorithm, called GABNI. The introduced method exploited an existing method, MIBNI, in the first stage to find an optimal solution and then GABNI if the first method fails due to the degree of complexity of an underlying regulatory function. Also in \cite{barman2020Boolean} {Barman and his colleagues} propose a pipeline for the gene regulatory network inference from time-series gene expression data by applying a statistical method called the chi-square method to infer a Boolean network from time-series gene expression data. They also suggested that structural accuracy can be increased by combining the chi-square test with neural networks as the perspective of their methodology. 

Applying different methods to automate this process may circumvent the limitation of a particular method and potentially improve the overall accuracy of the resulting BN~\cite{lee2009computational}. Besides benefiting from Boolean network inference from experimental data can also assist to evaluate and optimize the dynamic behaviour of the biological networks which provides a more reliable platform for further \emph{in silico} experiments and predictions~\cite{hartemink2000using}.

In this study, we present \taboon, an original work-flow for synthesizing BNs from biological data. This methodology utilizes biological information or experimental data in the form of the Boolean profiles for optimizing local formula inference and then Tabu-search algorithm to select the best BN candidate regarding significant biological features that have been defined for the system, viewed as global properties that the biological system complied.
These properties are usually used after the synthesis for validating the model. In our approach they are employed for the synthesis insuring that the generated model fulfils them by construction. 
 
After recalling the main features of the Boolean network (Section~\ref{sec:boolnet}) we detail the \taboon work-flow in Section~\ref{sec:taboon} which contains the description for Binarization~\ref{subsec:Binarization}, Local formula inference~\ref{subsec:local-inference} ,and the final synthesis~\ref{subsec:TaBooN sub}. In the \taboon benchmark (Section~\ref{sec:benchmark}) we present the application of the \taboon by using biological models befor concluding  (Section \ref{sec:Conclusion}) on the methodology and functionality of \taboon pipeline for automatic modelling and validating the biological BNs.

\section{Boolean Network}
\label{sec:boolnet}
Boolean network is a discrete dynamical system modelling the gene expression activity by capturing the functional transitions between two basic regulatory status: active or not defined by the Boolean values  $\Bool = \{1,0\}$ respectively.
Formally, it operates on Boolean variables $X$ by determining their \emph{state} evolution where a state $s$ is an interpretation assigning a Boolean value to the variables (\ie $s:X \to \Bool$). $\Bset_X$ denotes the set of all states for a set of variables $X$. A Boolean network is defined by a collection of Boolean functions, 
$ F= \{x_i = {f_{i}}(x_1, \ldots, x_n) \mid 1 \leq i \leq n\}, $ where each $f_i: \Bset_X \to \Bset_{x_i}$ is a propositional formula computing the state of $x_i$.

\paragraph{Model of dynamics.} The model of dynamics describes the state evolutions by a labelled transition system where the states are updated with respect to an updating policy, called the \emph{mode}. For example, in the \emph{asynchronous} mode, a single variable is updated per transition. Hence, the transition system is $\tuple{\evo, X, \Bool^n}$ where the transition relation $\evo \subseteq \Bset_X \times X \times \Bset_X$ is labelled by the updated variable such that\footnote{The complement  of a set $E$ by a subset $E' \subseteq E$ is noted $-E'= E \setminus E'$.}:$s \xevo{x_i} s' \eqdef s'=( f_i(s) \cup {s}_{-x_i}).$
Then, the global transition relation is defined as: $\evo = \bigcup_{x_i \in X} \xevo{x_i}$. 
 A path\footnote{ $\evo^*$ is the reflexive and transitive closure of the transition.} $s \evo^* s'$ characterizes a trajectory from $s$ to $s'$.
 
An \emph{equilibrium} $s$ is a particular state which is indefinitely reached once met \ie $\forall s' \in \Bset_X: s \evo^* s' \implies s' \evo^* s$. 
 A \emph{stable state} $s$ is a peculiar equilibrium satisfying the stability condition: $F(s)=s$. We denote by $\Omega_F$ the set of stable states of $F$ (\ie $\Omega_F = \{s \in \Bool_X \mid F(s)=s \}$ ).

\paragraph{Interaction graph} The \emph{interaction graph} $\mathcal{G}_F = \tuple{X, \arc}$ of network $F$ portrays the causal dependencies between the variables represented by the \emph{signed interactions} describing the nature of their regulatory activity. We denote the set of the regulators for target $x_j$ by $ \star \arc x_j$, and conversely $x_i \arc \star$ the set of targets of $x_i$.
An interaction $x_i \arc x_j$ exists whenever changing the value of $x_i$ may lead to a change in the value of $x_j$:
\begin{equation}
\begin{array}{l}
x_i \arc x_j \eqdef \\ 
\quad \exists s, s' \in \Bset_X: s_{x_i} \neq s'_{x_i} \land s_{-x_i}=s'_{-x_i} \land f_j(s)\neq f_j(s').
\end{array}
\label{eq:interaction}
\end{equation}

The \emph{signed interaction graph} $\tuple{Y, \arc, \sigma}$ refines the nature of the interactions by 
signing the arcs with the function $\sigma: (\arc) \to \{-1, 0, 1\}$ to represent a monotone relation between 
the source and target variables of the interaction~(\ref{eq:signed-interaction}); either increasing 
(label $1$, denoted $'+'$), or decreasing (label $-1$, denoted $'-'$), or neither (label $0$, denoted $'\pm'$), and formally defined as:
\begin{equation}
\begin{array}{l}
x_i \sarc{+} x_j \eqdef x_i \arc x_j \land \\
\; \forall s, s' \in\Bset_X: s_{x_i} \leq s'_{x_i} \land s_{-x_i} = s'_{-x_i} \implies f_{j}(s)\leq f_{j}(s') \\
x_i \sarc{-} x_j \eqdef x_i \arc x_j \land \\
\; \forall s, s' \in \Bset_X: s_{x_i} \leq s'_{x_i} \land s_{-x_i} = s'_{-x_i} \implies f_{j}(s)\geq f_{j}(s')
\end{array}
\label{eq:signed-interaction}
\end{equation}

\section{Boolean network synthesis}
\label{sec:taboon}
The algorithm infers a BN from binarized profiles of omic data. It is decomposed in two stages: the \emph{local formula inference} and the \emph{global network synthesis}. The former provides the set of formulas which is consistent with the data for each variable independently from the binarized expression profile and the regulatory graph. The latter selects among the resulting formulas the most appropriate ones in regards to some global properties to validate for the BN. These two stages act complementary by first finding all the consistent formulas compatible with the binarized expression profiles and by then generating a model resulting from an assembly of the formulas by selecting one per variable. Figure~\ref{fig:workflow} summarizes the different steps of the work-flow. 

\begin{figure}[h]
\begin{center}
\scalebox{0.9}{ 
\begin{tikzpicture}[auto]

\tikzstyle{block} = [rectangle, draw, fill=white, font=\sc, inner sep=3pt, 
text width=6.5em, text centered, rounded corners, minimum height=1.25cm, line width=1.5pt] 
\tikzstyle{inout}=[rounded rectangle, draw, text centered, fill=gray!20, minimum height=1cm, text width=8em, font= \footnotesize]
\tikzstyle{labeling}=[font=\it \small, black]
\tikzstyle{line} = [-latex, line width=3pt, white!45!black, 
text=black, text centered]
\tikzset{node distance=2.5cm}
\node [inout] (rnaseq) {Data};
\node [block, below of=rnaseq] (conversion) {Binarization};

\tikzset{node distance=4cm} 
\node [block, right of=conversion] (local) {Local \\ Inference};
\node [block, right of=local] (global) {Network Assembly};
\tikzset{node distance=2.5cm}
\node [inout, below of=global] (network) {Fittest\\ Boolean Network};
\node [inout, above of=local] (ig) {Interaction graph};
\node[inout, above of=global](constraints){Global properties};
\draw[line] (rnaseq)-- node[labeling, xshift=.3cm]{Normalized Data} (conversion);
\draw[line] (ig)-- (local);
\draw[line] (conversion) -- node[labeling, midway, yshift=-1.5cm]{Boolean profiles} (local);
\draw[line] (local) -- node[labeling, midway, yshift=-1.5cm]{Set of formulas/var.} (global);
\draw[line] (global)--(network);
\draw[line] (constraints)--(global);
\draw[line] (global) -- node[labeling, text width= 12ex, xshift=.3cm]{Boolean network} (network);
\end{tikzpicture}
}
\end{center}
\caption{\taboon work-flow}
\label{fig:workflow}
\end{figure}

\subsection{Binarization}
\label{subsec:Binarization}
The binarization or Booleanization consists in deducing Boolean profiles from quantified gene expression data. The discretization process can be understood as a partitioning of gene expression value into functional classes respectively representing whether a gene is activated ($1$) or inhibited ($0$) that are defined by the value of their expression. Hence the binarization basically corresponds to a bi-partitioning approach where the two classes describe the regulatory status: active or inactive. However, a third class is usually added for collecting gene expression value with status remaining undetermined during analysis. The computational methods differ by the nature of the analysed data: time series~\cite{Hopfensitz2012, Muessel2016}, differential expression~\cite{Anders2010, Robinson2010}, or pseudo-global based on data coming from a set of different conditions~\cite{Jung2017, Beal2018}. The core of the methods lies on the discovery of a threshold delineating the gene status (active or inactive). We assume that this part is tackled by the above methods and the input of the work-flow is a set of binarized expression profiles. 

\subsection{Local formula inference}
\label{subsec:local-inference}
For the local inference, the Boolean profiles are assimilated to rows of a Boolean truth table where the profiles of regulators correspond to the input profiles and the regulated variable to the output profiles of the truth table. As example if we do assume that $x_2, x_3, x_4$ regulate $x_1$ and we obtain the following profiles $P=\{ (1, 1, 0, 1), (0, 1, 1, 0), (1, 0, \_ , 0)\}$, we do consider the following partial truth table (\ref{eq:truth-table}) for their interpretation:
\begin{equation}
	\begin{array}{|c c c | c|}
		\hline
		x_2 & x_3 & x_4 & x_1 \\
		\hline
		1 & 0& 1& 1 \\
		 1& 1& 0& 0\\
		 0& \_& 0 &1\\
		 \hline
		\end{array}
\label{eq:truth-table}
\end{equation}

Hence, compared to a full  truth table, the resulting table is partially defined and possibly with missing values (\ie $\_$ sign in Table~\ref{eq:truth-table}). The local formula inference method will complete the truth table by characterizing all the outputs. This process lies on a set of constraints related to the regulation for discovering all the possible outputs completing the partial truth table from which the formulas consistent with the binarized profiles and the regulation can be deduced.
The inference method is thus seen as a satisfiability problem on propositional formula~\cite{Garey2002} instantiating the output values of the truth table. The issue is thus to define the appropriate set of constraints for the outputs assimilated to the variables of the satisfiability problem. 

By convention, the variables representing the outputs of a truth table are denoted by $b_p$ and the label $p$ is the input Boolean profiles. For example given the input profile $(1, 0, 1)$ the variable representing the output value is denoted $b_{101}$. $x_1$ will always be the output variables while the others ($x_i, i >1$) are the input variables by convention. 
Hence, a Boolean profile $p$ of the regulators yielding to value $v$ ($p(x_2, \ldots, x_n)=v$) for the target $x_1$ is represented by the equation $b_p=v$ formulated in propositional logic as an equivalence: $b_p \iff v $.

Two kinds of constraints are considered: the definition of the profiles as formulas characterizing the \emph{Boolean profile based constraint}, and the constraints strictly related to the regulation characterizing the \emph{regulation based constraints}.
 
\subsubsection*{Boolean profile based constraint} 
The Boolean profiles are formulated as logical equivalences (\ie $b_p \iff v$) while however taking the variables with undetermined values into account. We thus need to formalize the equivalences as a combination of defined and undefined values. Recall that when $v$ is known, the equivalences are simplified according to the following rules $b \iff 1$ equates $b$ and $b \iff 0$ equates $\neg b$. Then the equivalence is concretely never explicitly formulated in the constraint related to the profiles while being used for the constraint specification.

Now, assume that the variables with an undetermined value are the variables before rank $m$, $x_2, \ldots, x_m$, and the values for the other variables are determined, $x_{m+1}, \ldots, x_r$, we state that at least one variable of the form $b_{\star, p(x_{m+1})\ldots p(x_r)}$ equals $p(x_1)$ for a partial Boolean profile $p$. This statement is formalized as a disjunction over the valuation of the undefined variables:
 \begin{equation}
 \mathcal{C}^{\textit{bp}}_p=\bigvee_{ \beta \in \Bool^{m-1}} \left(b_{\beta p(x_{m+1})\ldots p(x_r) } \iff p(x_1)\right).
 \label{eq:constraint1}
\end{equation}

\subsubsection*{Regulation based constraints}
Two kinds of constraints are deduced from the regulation: the \emph{consensus regulatory profile} and the \emph{regulation conformity} originated from the definition of the interactions.

\paragraph{Consensus regulatory profile.} This constraint corresponds to a profile where all the regulators cooperate to set the target to a specific value. Two cases are considered: 
\begin{itemize}
\item all the inhibitors are \textsc{off} ($0$) and all the activators are \textsc{on} ($1$), then the target is necessary switched \textsc{on} according to the definition of the regulation.
\item Conversely, when all the inhibitors are \textsc{on} ($1$) and all the activators are \textsc{off} ($0$), then target is switched \textsc{off} by definition.
\end{itemize}
We use the labelling convention on $b_\star$ variables to express them. Assume that the inhibitors correspond to the variables $R^-=x_{2}, \ldots, x_k$ and $R^+=x_{k+1}, \ldots, x_r$ are the activators, the constraints are defined by the following formula:

\begin{equation}
 \mathcal{C}^\textit{cp}_{R^-, R^+} = b_{\underbracket{0\ldots0}_{k-1}\underbracket{1\ldots1}_{r-k}} \land \neg b_{\underbracket{1\ldots1}_{k-1}\underbracket{0\ldots0}_{r-k}}
 \label{eq:constraint2}
\end{equation}
When some regulations remain unknown this constraint is extended similarly to (\ref{eq:constraint1}) by a disjunction applying all the Boolean configurations for the remaining unknown regulator fulfilling Constraint~(\ref{eq:constraint2}).

\paragraph{Regulation conformity.} This constraint characterizes the fact that the values of $b_\star$ variables should also comply to the positive and negative regulation rules (\ref{eq:signed-interaction}). This constraint is applied to all regulators independently and expresses the state order characterizing the regulation constraint on the order on labels for $b_\star$ variables.

Assume that the regulators of $x_1$ are $x_2, \ldots, x_r$, the positive interaction $x_2 \signed{+} x_1$ leads to the following constraint:
$$
\begin{array}{l}
\exists s_1, s_2 : 
\\ \quad s_1(x_2) < s_2(x_2) \land s_{1_{-x_2}}=s_{2_{-x_2}} \land f_1(s_1) < f_2(s_2) \quad \land \\
\forall s_1, s_2 : 
\\ \quad s_1(x_2) < s_2(x_2) \land s_{1_{-x_2}}=s_{2_{-x_2}} \implies f_1(s_1) \leq f_2(s_2).
\end{array}
$$

This condition will be formulated in propositional logic with $b_{\star}$ by using the order on the labels of variables. The sole configuration satisfying 
$s_1(x_2) < s_2(x_2)$ is $s_1(x_2)=0, s_2(x_2)=1$. The positive regulation rule must be checked for this configuration only since it holds for the others.
 By scanning all the remaining configurations stating the equality $s_{1_{-x_2}}=s_{2_{-x_2}}=\alpha, \alpha \in \Bool^{r-2}$, and as $f_1(s_1)= b_{0\alpha}$ and $f_1(s_2)= b_{1\alpha}$, the positive regulation rule is equivalently expressed with $b_\star$ variables as:
$$ \left(\exists \alpha \in \Bool^{r-2}: b_{0\alpha} < b_{1\alpha} \right) \land \left(\forall \alpha \in \Bool^{r-2}: b_{0\alpha} \leq b_{1\alpha}\right). $$
As the set of regulators is finite, the quantifiers $\exists, \forall$ can be respectively expressed by $\bigvee_{\alpha \in \Bool^{r-2}} , \bigwedge_{\alpha \in \Bool^{r-2}} $, and the orders are translated into equivalent formulas: $\neg x \lor y$ for $x \leq y$ and $\neg x \land y$ for $x < y$. 
Hence, the condition is formulated in propositional logic as follows: 
\begin{equation}\mathcal{C}^{r+}_{x_2} =
 \bigvee_{\alpha \in \Bool^{r-2}} (\neg b_{0\alpha} \land b_{1\alpha}) \quad \land 
\bigwedge_{\alpha \in \Bool^{r-2}} (\neg b_{0\alpha} \lor b_{1\alpha}).
 \label{eq:constraint3}
\end{equation}
Following the same reasoning, the constraint for the negative regulation $x_2 \signed{-} x_1$ is:
\begin{equation}\mathcal{C}^{r-}_{x_2} =
 \bigvee_{\alpha\in \Bool^{r-2}} ( b_{0\alpha} \land \neg b_{1\alpha}) \quad \land \bigwedge_{\alpha \in \Bool^{r-2}} ( b_{0\alpha} \lor \neg b_{1\alpha}).
 \label{eq:constraint4}
\end{equation}

\subsubsection*{Constraint assembly and formula inference}
The final constraint is composed of a conjunction of the previous constraints.
Table~\ref{tab:example} reports the different constraints used by the inference for the example
given introduction of the section (Truth table~\ref{eq:truth-table}) such that $x_2$ is a positive regulator, $x_3$ a negative regulator and the regulation of $x_4$ is unknown, that is: 
$x_2 \signed{+} x_1, x_3 \signed{-} x_1, x_4 \arc x_1.$

\begin{table}[ht]
\begin{center}
\tabulinesep=_5pt^5pt
\begin{tabu} to \textwidth {X[1, $$l]} %
\tabucline-
 \mathcal{C}^\mathit{bp}_{P(x_1)} =b_{101}\land \neg b_{110}\land \left(b_{000}\lor b_{010}\right). \\
\mathcal{C}^\textit{cp}_{R^-(x_1), R^+(x_1)}= \left(b_{100}\land \neg b_{010}\right)\lor \left(b_{101}\land \neg b_{011}\right).\\

 \mathcal{C}^{r+}_{R^+(x_1)}=\left(\left(\neg b_{000}\land b_{100}\right)\lor \left(\neg b_{001}\land b_{101}\right)\lor \left(\neg b_{010}\land b_{110}\right)\lor \left(\neg b_{011}\land
 b_{111}\right)\right)\land \left(\neg b_{000}\lor b_{100}\right)\land \left(\neg b_{001}\lor b_{101}\right)\land \left(\neg b_{010}\lor b_{110}\right)\land  \left(\neg b_{011}\lor b_{111}\right) \\
 
\mathcal{C}^{r-}_{R^-(x_1)} = \left(\left(b_{000}\land \neg b_{010}\right)\lor \left(b_{001}\land \neg b_{011}\right)\lor \left(b_{100}\land \neg b_{110}\right)\lor \left(b_{101}\land \neg
 b_{111}\right)\right)\land \left(b_{000}\lor \neg b_{010}\right)\land \left(b_{001}\lor \neg b_{011}\right)\land \left(b_{100}\lor \neg b_{110}\right)\land
 \left(b_{101}\lor \neg b_{111}\right)\\
\tabucline-
\end{tabu}
\end{center}
\caption{Constraints related to the example.}
\label{tab:example}
\end{table}

Finally, the resolution of the global constraint is performed by a SAT-solver instantiating all the $b_\star$ variables. Each particular instantiation of the variables thus provides the result of the truth table for each entry. Once the truth table is completed, the minimal disjunctive normal form of a formula is straightforwardly deduced using prime implicant method~\cite{McCluskeyJr1956} to generate the resulting formula. The different instances of the $b_\star$ variables define the set of formulas related to the analysed target variable ($x_1$ here).
For the example, three formulas are found~(Table~\ref{tab:example}):

\medskip
	\begin{tabu} to 0.9\textwidth {X[l,$$]}
		\left(x_2\land x_4\right)\lor \neg x_3, \\ 
		\left(x_2\land \neg x_3\right)\lor \left(x_2\land x_4\right)\lor \left(\neg x_3\land \neg x_4\right), \\
		\left(x_2\land \neg x_3\right)\lor \left(\neg x_3\land \neg x_4\right).
	\end{tabu}

If the regulation of $x_4$ is known, for example $x_4 \signed{+} x_1$, the resolution returns the first formula only.

\subsubsection*{Experimental evaluation} 

\begin{figure}[htb]
\begin{center}
\includegraphics[width=0.9\textwidth]{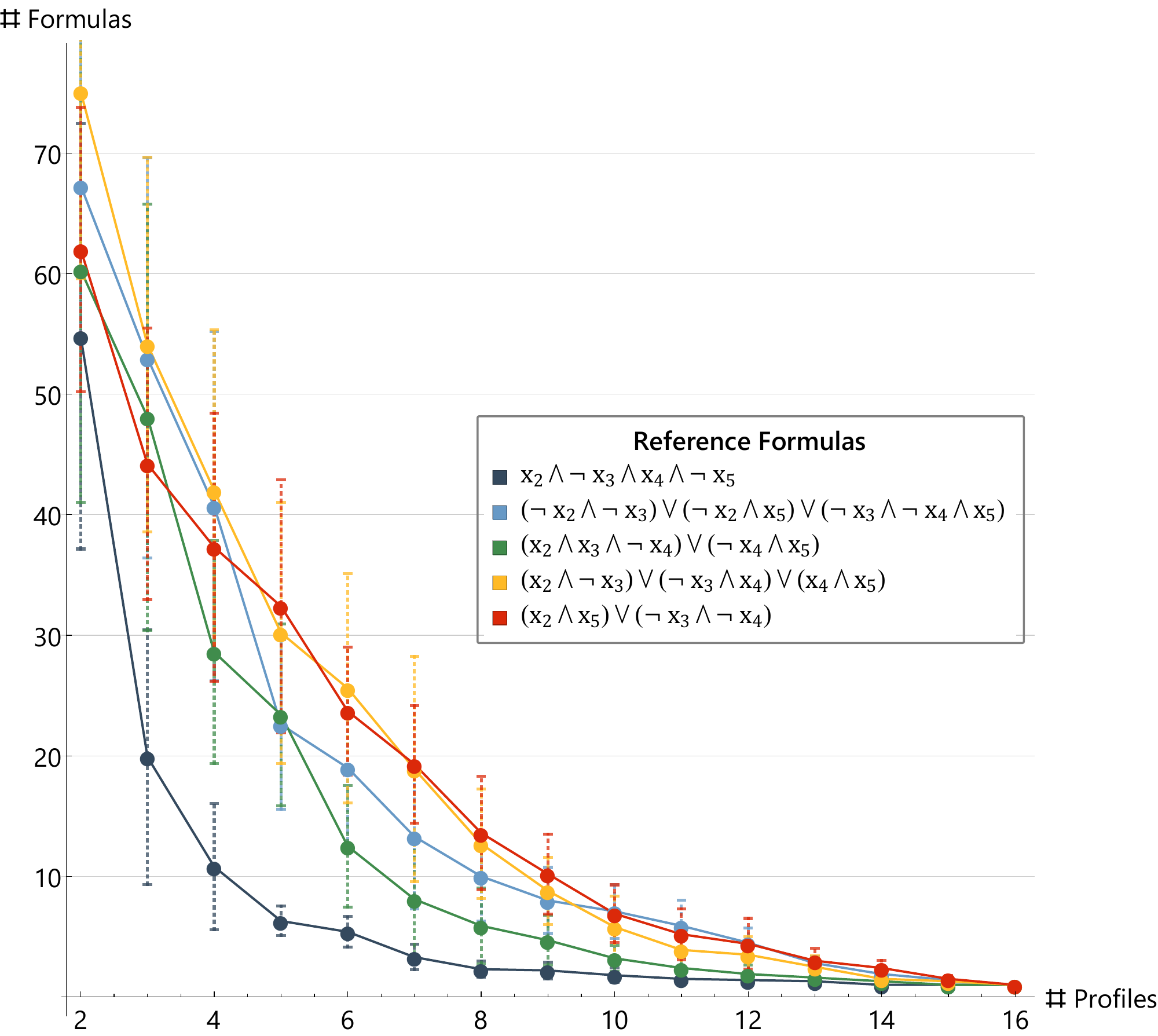}

\medskip
\begin{minipage}{0.9\textwidth}
 \footnotesize
The curves depict the average number of the inferred formulas from $10$ trials of profile sets randomly selected among the lines of the truth table for $5$ reference formulas. The error-bars show the standard deviation. All the regulators are defined.
\end{minipage}
\end{center}
\caption{Evaluation of the local inference formula algorithm.}
\label{fig:experiment}
\end{figure}

The accuracy of the method is experimentally evaluated on predefined formulas that the method attempts to discover with Boolean profiles randomly picked from their truth table. Less formulas are inferred more accurate the result is. Figure~\ref{fig:experiment} describes the average number of inferred formulas on $10$ trials for $5$ different reference formulas\footnote{Extension of the experiments on a larger set of formulas and trials (not reported here) lead to the same the conclusions.} involving $4$ variables. The number of profiles gradually increases by adding a new profile not already selected to the previous ones at each step. 

The method shows a high accuracy in comparison to the number of possible formulas which is $2^{2^n}$ for $n$ variables. Indeed, for $4$ variables the number of possible formulas is $65\, 536$ and the maximal number of inferred formulas never exceed $100$. Thus, the inferred formulas represents less than $\approx 0.15\%$ of the possible formulas. However, this minute number could remain large as the number of possible formulas is super exponential. The error bars showing the standard deviation reveal a significant sensitivity of the method to the chosen profiles since their number is the same. The means and the standard deviations have an exponential decay stressing the importance of the amount of Boolean profiles provided for the resolution. With a complete set of profile only the tested formula is found. The number of formulas also seemingly depends on the number of terms and clauses. For example, the number of potential formulas for the first reference formula ($x_2\land\neg x_3 \land x_4 \land \neg x_5$) with the least number of terms is significantly lower than the others. The complexity of the algorithm is in $\mathcal{O}(2^{2^n})$ since SAT problem is NP-Complete with $2^n$ variables representing the outputs of the truth table. However this upper bound does not truly represent the empirical performance of the method which is efficient in practice for $n \lesssim 10$ variables with a sufficient amount of Boolean profiles. Moreover, the inference can be processed in parallel for a network as the inference is performed independently on each variable.

\subsection{TaBooN}
\label{subsec:TaBooN sub}

The final inferred network results from an assembly of the found formulas by selecting one formula per variable. Thus, the number of potential network corresponds to the product of the number of formulas related to variables that may be huge (Figure~\ref{fig:benchmark-nb-models}).
The selection of a network must therefore be carried out sparingly by selecting the formulas appropriately for yielding a truthful model. This issue is seen as an optimization problem where the putative networks are assessed by quantifying the ``truthfulness'' with an objective function. \taboon method uses the Tabu Search~\cite{Glover1986, Glover1998} that offers a suitable framework for encoding the network inference. 

Tabu-search is a meta-heuristic superimposed on another heuristic to explore the best alternative solutions by moving in the neighbourhood of the last found one. 
Adaptive memory-based strategies are the hallmark of tabu search approaches to circumvent the entrapment in local minima. 
A memory structure storing the Tabu moves prevents to repeatedly visit the same moves consecutively in order to avoid staying in local minima. Mid and long term memories are respectively used for the intensification around the elite solutions or the diversification opening to alternative solutions. 

For the inference of model, the search space represents all the Boolean networks and two Boolean networks are neighbours if and only if they differ in one formula. Therefore a move consists in changing one formula for one variable. To specialize the Tabu search for network inference, we need to define the objective function scoring the truthfulness of a model and the local heuristic search procedure determining the best local move from the current network. 

\subsubsection*{Objective function} 
 The objective function quantifies the truthfulness of a network model based on properties used to validate model compared to the feature of the studied biological system or function. 
 Although no specific rules govern the definition of the objective function, some patterns related to the validation principle of a Boolean network drives its definition. 
 
Frequently, a Boolean network is validated by the Boolean signatures defining the molecular states of the phenotypes for some biomarkers~\cite{Mendoza1999, Sahin2009, Traynard2016, Verlingue2016, Cohen2015, Enciso2016}. These signatures are expected to be met at stable states in a model because the reproducibility of the phenotypes is modelled by the stability condition, thus assimilating the signatures as a part of the stable states. 
Notice, that the determination of the stable states can be processed efficiently by a symbolical computation using SAT solver~\cite{Dubrova2011}. 

Another criterium concerns the monotony of the network. A monotone model contains only pure activations and inhibitions providing a clean model reducing its complexity by limiting the non-monotone interactions. 
 
A third criterium assessing the specificity of a model is the number of stable states
that should corresponds to the number of the desired phenotypes controlled by the network. 
Indeed, It is assumed in modelling that each phenotype must correspond to an equilibrium. Hence, models with a large number of stable states are considered as a coarse specification of the biological function because many phenotypic alternatives are carried out by this network. Therefore, the underlying dynamics does not accurately reflect the studied function.
 
 Hence, several criteria are considered for selecting the fittest network leading a multi-criteria function objective.Therefore the comparison between scores is achieved by the minimum of Pareto. 

\subsubsection*{Move procedure}

 \begin{figure}[h]
 \centering
 \includegraphics[width=0.85\textwidth]{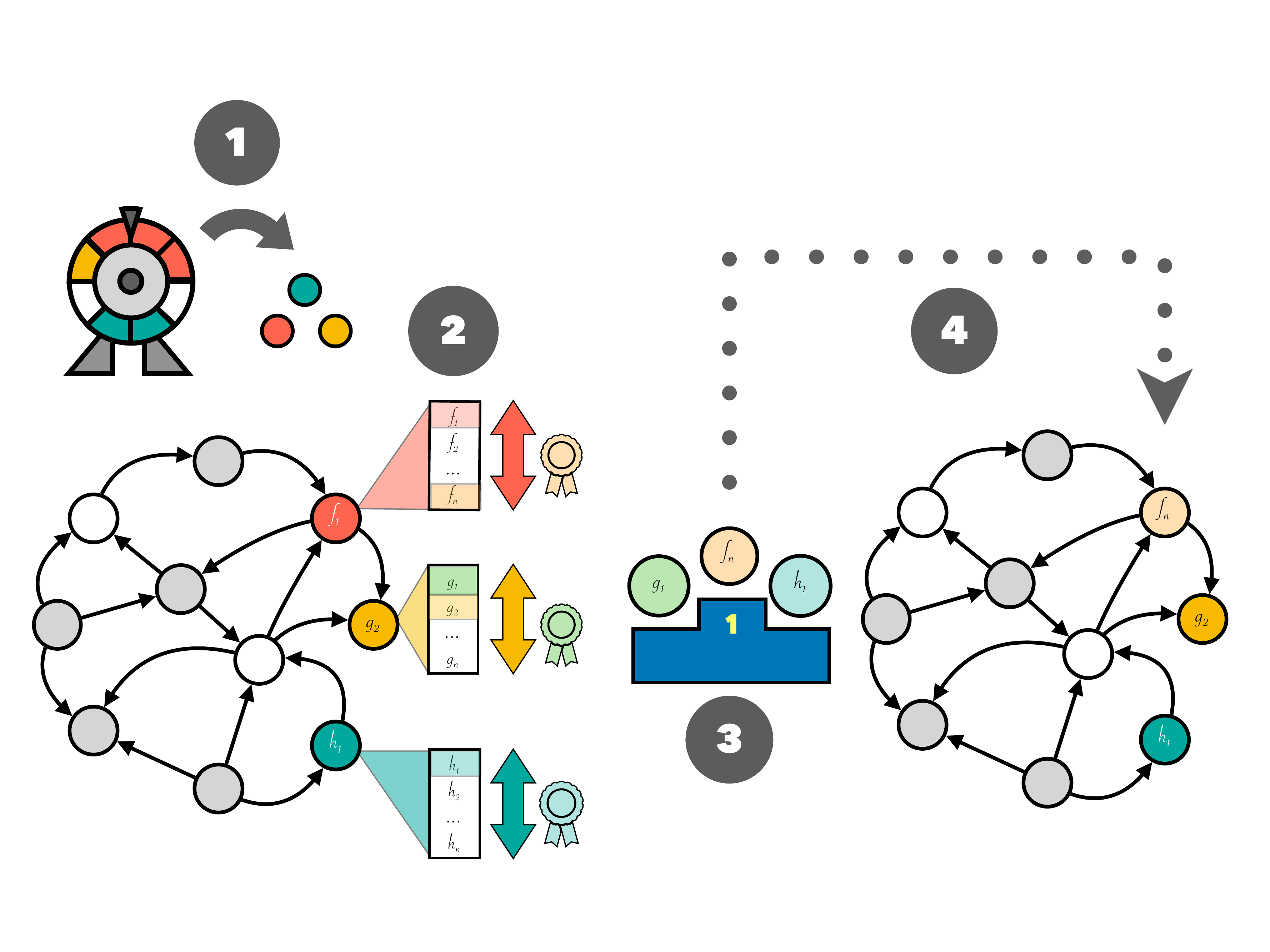}
\begin{minipage}{0.9\textwidth}
\footnotesize
The steps leading to a move are:
\begin{enumerate}
\item Select candidates outside the Tabu list (grey nodes) using a biased wheel based on centrality measure.
\item For each candidate select the local best formula by iteratively replacing the initial formula with all the formulas of the candidate and then compare the score of the resulting networks for finding a solution which is Pareto minimal.
\item Elect the move corresponding to the best formula among the local best formulas associated to a candidate.
\item Replace the initial formula by the best formula for the candidate. 
\end{enumerate}
\end{minipage}
 \caption{Computational steps of a move.}
 \label{fig:taboon}
 \end{figure} 

The local move consists in replacing one formula by another for the current Boolean network. 
The local search procedure is based on an heuristic selecting the best formula for a network among the sets of formulas computed during the local formula inference phase (Sub-Section~\ref{subsec:local-inference}). Due to the potential large number of formulas, we decompose the move in two steps: first some variable are chosen as candidates for the formula change, and then the best formula is elected from the set of formulas of these candidates. The number of candidates at each step is a parameter of the model. Figure~\ref{fig:taboon} summarizes the steps leading to a move.

\paragraph{Move on candidates} The impact of the modification of the formula differ among candidate depends on the regulation capacity. Therefore, the selection of a candidate is based on a ranking according to their influence in the network. A change of the formula with the highest ranked variable would likely be more consequential on the dynamics than a subsidiary one.
Centrality analysis has been widely used to find influential nodes in networks, with a large spectrum of applications in systems biology analysis~\cite{ashtiani2018}. The variable ranking is computed from a centrality measure. However, we use the biased roulette wheel method~\cite{Srinivas1994} where the size of wheel sections for the variables is proportional to the centrality measurement for privileging the variables with a high centrality while still being able to elect any variable. The probability of choice for a variable with a centrality measure $c_i$ is thus:
$\frac{c_i}{\sum_{x_j \in X'} c_j}$ where $X' \subseteq X$ is the set of current available variables outside the tabu list and not already chosen as candidate.
We use the Eigenvector centrality measuring the influence of a node in a network
for the experiments (Section~\ref{sec:benchmark}). 
\paragraph{Move on formulas} Once the candidates are selected for a move, the scores are computed by exchanging the formula for each candidate with all their alternative formulas. Then, the best local formula which is a Pareto minimum is chosen. Finally a move corresponds to the selection of the formula with the best score elected from the best local formulas selected for each candidate. Then, only one formula changes for one candidate during a move. 
 
\subsubsection*{Initialization, diversification and halting condition}
The best network is updated if the score of the current network minimizes it. The halting condition is based on two conditions: A bound on the number of consecutive failures, and reaching the minimal score which is a null vector. The diversification is used for favouring the visit of all the variables. This strategy is motivated by the necessity to adjust all the local dynamics of the variables because they all contribute to get the expected global property. The long term memory stores the frequency of the moves and choose one move which is less frequently visited at each iteration. Finally, the initial network can be designed by a modeller or randomly by selecting one formula per variables. To our knowledge, the quality of the first network speeds the convergence to the best network but does not seem to alter the quality of the final result. Notice that this method may thus be complementary used for improving already designed BNs viewed as the initial network.
 
\section{TaBooN benchmark} 
\label{sec:benchmark}

The benchmark\footnote{ The parameters used for the experiments are: number of moves= $4$, maximal number of formulas = $500$, failure bound = $100$.} evaluates the truthfulness of a network obtained by \taboon with regards to the availability of data (\ie Boolean profiles) and the objective function.
To enable a comparative analysis, we evaluate the \taboon method on five published Boolean networks\footnote{By convention, they will be named by the initial of the three first authors name for their identification.} modelling different biological processes. We consider them as the \emph{references} meaning that they are optimal solutions. Under this hypothesis the truthfulness of a network is assessed by its closeness to a reference network. The main characteristics of the reference networks are summarized in Table~\ref{tab:benchmark-networks}.

 \begin{table}[ht]
\centering
\tabulinesep =_1ex
\begin{tabu} to 0.95\textwidth {X[1, c] X[5, l] X[1, c]}
\tabucline[1.5pt]{} 
Name & \multicolumn{1}{c}{Modelled biological process} & Reference \\
\tabucline[1pt] {}
 \textsc{sfl} & 
\textsc{erb} \textsc{emm} receptor regulated \textsc{g1/s} transition network used for anticancer drugs analysis. 
& \cite{Sahin2009} \\
\textsc{tff} &
Network of Mammalian cell cycle& \cite{Traynard2016}\\
 \textsc{vds}& Signalling network controlling the S-phase entry and 
 geroconversion senescence. & \cite{Verlingue2016} \\
 \textsc{cmr} & Regulatory network describing Epithelial to Mesenchymal Transition mechanism & \cite{Cohen2015} \\
 \textsc{emm} & Network interconnecting the communication pathways between haematopoietic stem cells and mesenchymal stromal cells & \cite{Enciso2016}\\
\tabucline[1.5pt]{}
\end{tabu}

\medskip
\begin{equation*}
\begin{array}{|c|cc|cc|cc|}
\hline
 \text{Name of} & 
 \multicolumn{2}{c|}{\text{General}} &
 \multicolumn{2}{c|}{\text{Mean degree}} & \multicolumn{2}{c|}{\text{Max degree}} \\
 \text{Model} & \text{$\#$ vertices.} & \text{$\#$ edges} & \text{In} & \text{Out} & \text{In} & \text{Out} \\
 \hline
 \textsc{sfl} & 20 & 51 & 2.55 & 2.55 & 5 & 6 \\
 \textsc{tff} & 11 & 39 & 3.55 & 3.55 & 6 & 6 \\
 \textsc{vds} & 25 & 67 & 2.68 & 2.68 & 6 & 8 \\
 \textsc{cmr} & 32 & 157 & 4.91 & 4.91 & 8 & 13 \\
 \textsc{emm} & 26 & 81 & 3.12 & 3.12 & 9 & 7 \\
 \hline
\end{array}
\end{equation*}
\caption{Properties of the reference networks.}
\label{tab:benchmark-networks}
\end{table}
The solution networks are compared to the reference networks by using a distance representing the percent of dissemblances on the result of the evolution functions. 
The \emph{truth-value distance} $d_T$ (\ref{eq:truth-distance}) counts the number of differences of the outputs of the truth table for the formulas having the same number of variables with regards to all the possible inputs. Let $f, f'$ be two formulas with the same number of variables ($n$), the truth-value distance is defined as:
\begin{equation}
 d_{T}(f, f')= \frac{\sum_{s \in \Bool^n} d_H(f(s), f'(s))}{2^{n}}; 
\label{eq:truth-distance}
\end{equation}
where $d_H$ is the Hamming distance.
For example, $ d_T( x \lor y, x \land y)=0.5 \; (50 \%)$. Indeed by considering $00, 01, 10, 11$ as the sequence of inputs for both formulas the outputs are respectively $(0, 1, 1, 1)$ for \textsc{or} and $(0, 0, 0, 1)$ for \textsc{and}. Thus the formulas differ for half of the inputs: $01, 10$. 

The extension of this distance on networks with the same interaction graph corresponds to the mean of the truth value distances between pairs of formulas defining the evolution of the same variables in each network: 
\begin{equation*}
 d_T(F, F')=\operatorname{mean} {\{ d_T(f_i, f'_i) \mid f_i \in F, f'_i \in F'\}}.
\end{equation*}

The objective function is defined on the three criteria previously mentioned: the occurs of signatures at stable state, the monotony of the model and the number of stable states that should equal the number of stable states of the reference network ($\phi$). An optimal score corresponds to a null vector.
 $\Gamma_B = \{ \hat s_B \mid \hat s \in \Omega_\phi \}$ defines the set of signatures for the biomarkers. Each signature ($\hat s_B$) must be included in one stable state for an optimal Boolean network $F$. 
 
The objective function for a Boolean network $F$ returns a vector of scores formally defined in (\ref{eq:score-fun}) where $\dot \cup$ stands for the concatenation operator. 
The score of a signature is assessed by the minimum of the Hamming distance ($d_H$) of the stable states to a signature. It is separately applied to each signature providing an independent score for each. Notice that when the signature is included in a stable state, the score is $0$ otherwise the function returns the distance of the closest equilibrium to the signature. The monotony is evaluated by counting the number of non-monotone interactions. The score of a monotone network is thus $0$. Finally the score related to stable states is the absolute difference between the number of stable states of $F$ and number of those of the reference network $\phi$.
\begin{multline}
\overbracket{\dot \bigcup_{\hat s_B \in \, \Gamma_B}
	\min \{ d_H(\hat s_B, s_B) \mid s_B \in \Omega_F \}
}^\text{Biomarker signatures} 
\dot \cup 
 \overbracket{{\sum_{ f_i \in F} \mathbf{1}_{ \mathcal{\bar M}} (f_i))}^{\vphantom{X}} }^\text{Monotony} 
 \dot \cup \overbracket{ 
	{\operatorname{abs} (|\Omega_F| - |\Omega_\phi|)}^{{\vphantom{X}^{\vphantom{X}}}^{\vphantom{X}}}}^\text{Stable states}, 
\label{eq:score-fun} 
\end{multline} 
where $\mathbf{1}_{\mathcal{\bar M}}$ is the indicator function of the non-monotone functions (\ie $\mathbf{1}_\mathcal{\bar M}(f)=$ $0$ if $f$ is monotone and $1$ if not). The number of used criteria is thus $|\Gamma_B|+2$.

The goal is to assess the impact of the variation of two parameters: the amount of Boolean profiles representing the binarization of the \textsc{rna}-seq expression, and the number of biomarkers used for the signatures. We evaluate how the closeness based on the truth-value distance to a reference network defined in Table~\ref{tab:benchmark-networks} evolves. 

 For each network, a given percent of the entries of the truth table is randomly selected and assimilated to the input binary profiles for the formula inference (Sub-Section~\ref{subsec:local-inference}). A new set of formulas is computed whenever the profile percent varies.
As the biomarkers presumably represent relevant molecules for the study, we order the genes according to a priority stressing their influence determined from their Eigenvector centrality and we incrementally choose a growing number of the biomarkers following this order. The signatures correspond to the values of the chosen biomarkers for each stable state of a reference network. $10$ trials are performed for each percent value of binary profiles and signature size.
 For Boolean profile variation, the size of the biomarkers is fixed to $25\%$ of the number of variables. For signature variation, the percent of binary profiles is fixed to $50\%$ and the formulas correspond to those computed for this percent. 
 
 \begin{figure}[h]
 	\centering
 	\includegraphics[width=0.9\textwidth]{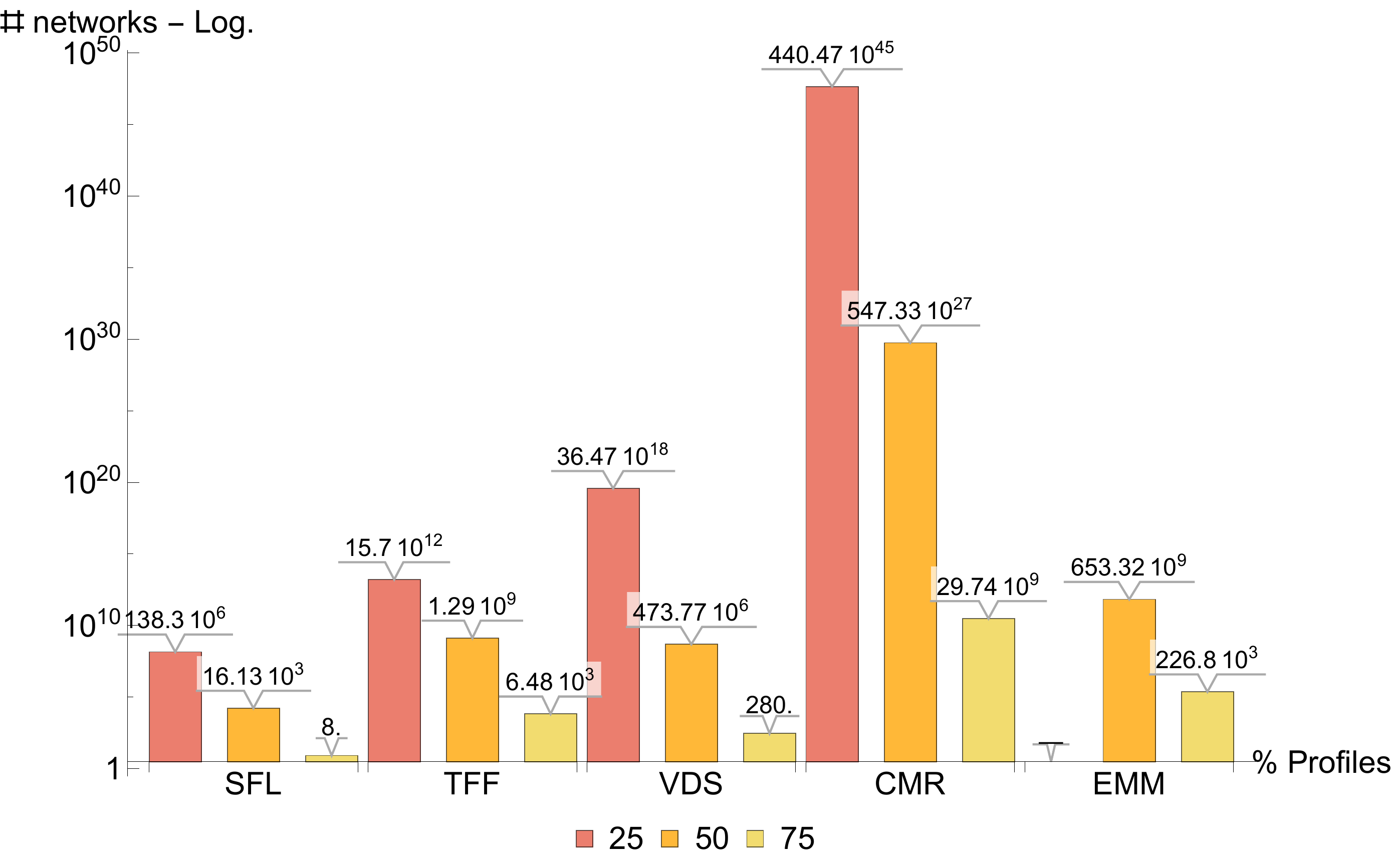} 
 	
 	\begin{minipage}[t]{0.9\textwidth}
 		\footnotesize
 		The bar chart reports the number of potential models (\ie model space cardinality) from which a solution is selected by \taboon.
 	\end{minipage}
 	\caption{Number of potential models.}
 	\label{fig:benchmark-nb-models}
 \end{figure}

\begin{figure}[p]
	\centering
	\extrarowsep=0.5ex
	\begin{tabu} to \textwidth {X[c] X[c]}
		\textsl{ 1) Truth-value distance.} &\textsl{ 2) Computation time.} \\
		\multicolumn{2}{c}{\rule[0.5ex]{0.43\textwidth}{0.75pt} \hfill \textsc{Profile} \hfill \rule[0.5ex]{0.43\textwidth}{0.75pt}} \\
		\includegraphics[width=0.45\textwidth, valign=b]{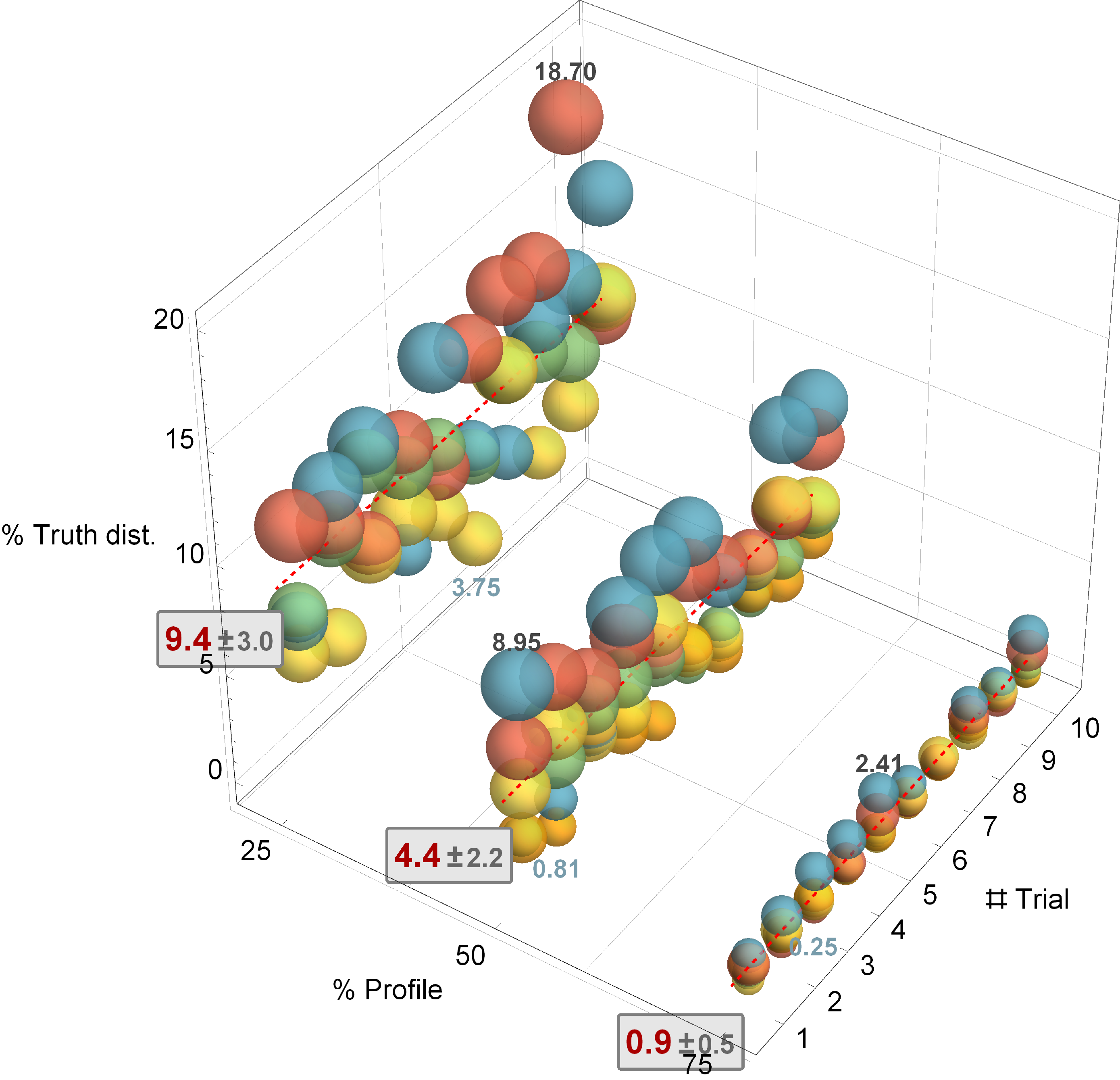} & 	
		\includegraphics[width=0.45\textwidth, valign=b]{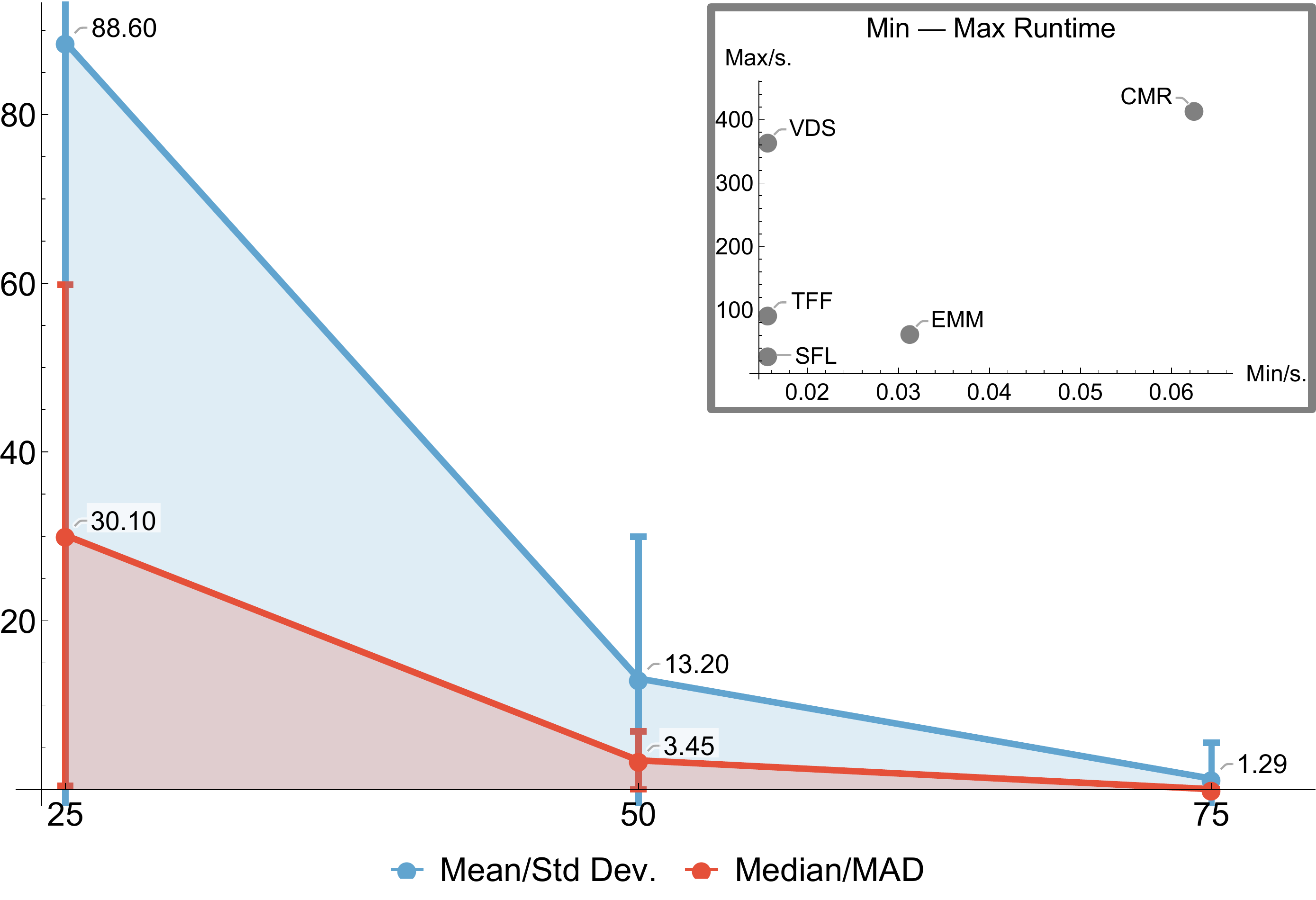} \\
		\multicolumn{2}{c}{\rule[0.5ex]{0.43\textwidth}{0.75pt} \hfill \textsc{Signature} \hfill \rule[0.5ex]{0.43\textwidth}{0.75pt}} \\
		\includegraphics[width=0.45\textwidth, valign=b]{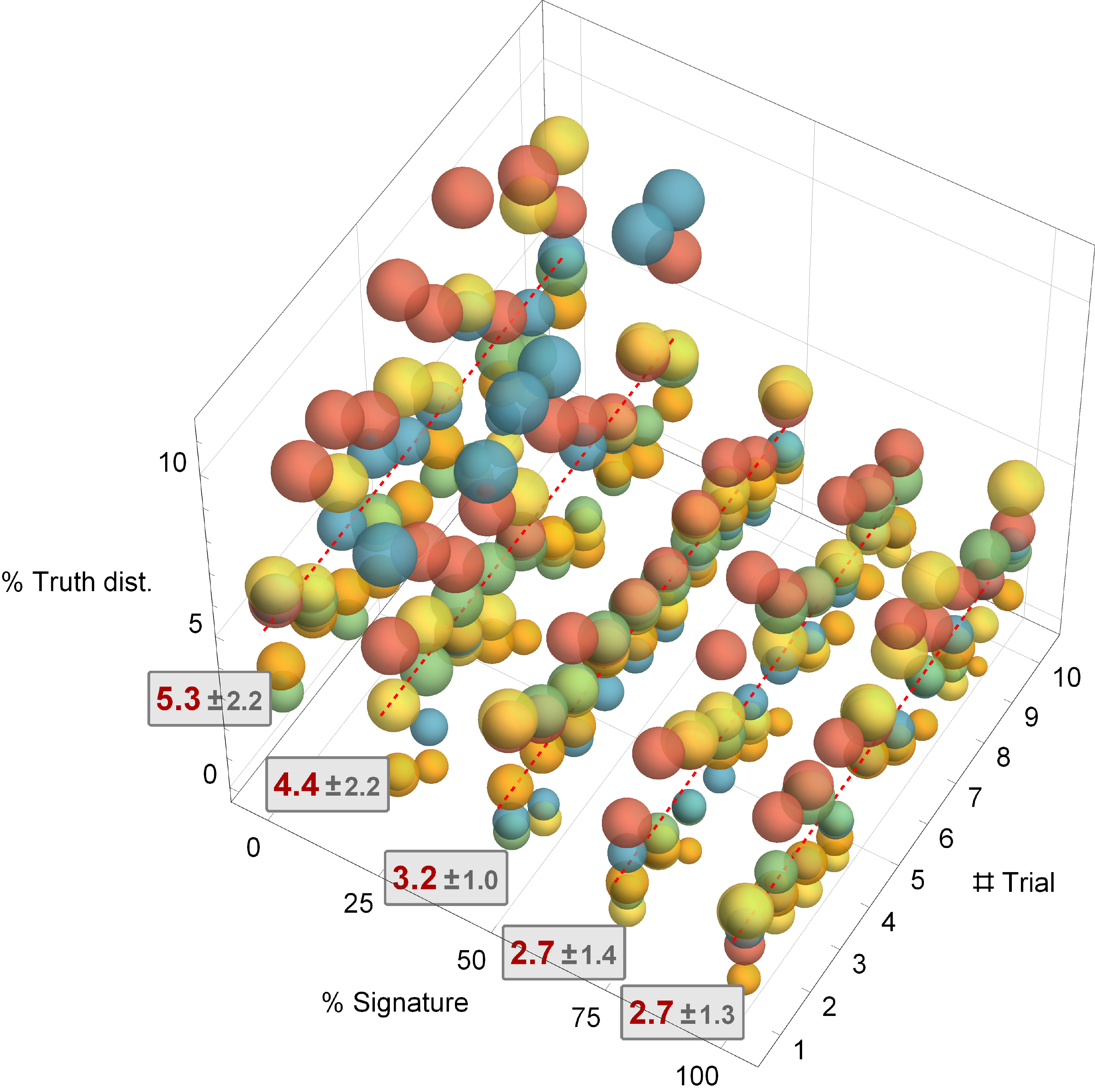} &
		
		\includegraphics[width=0.45\textwidth, valign=b]{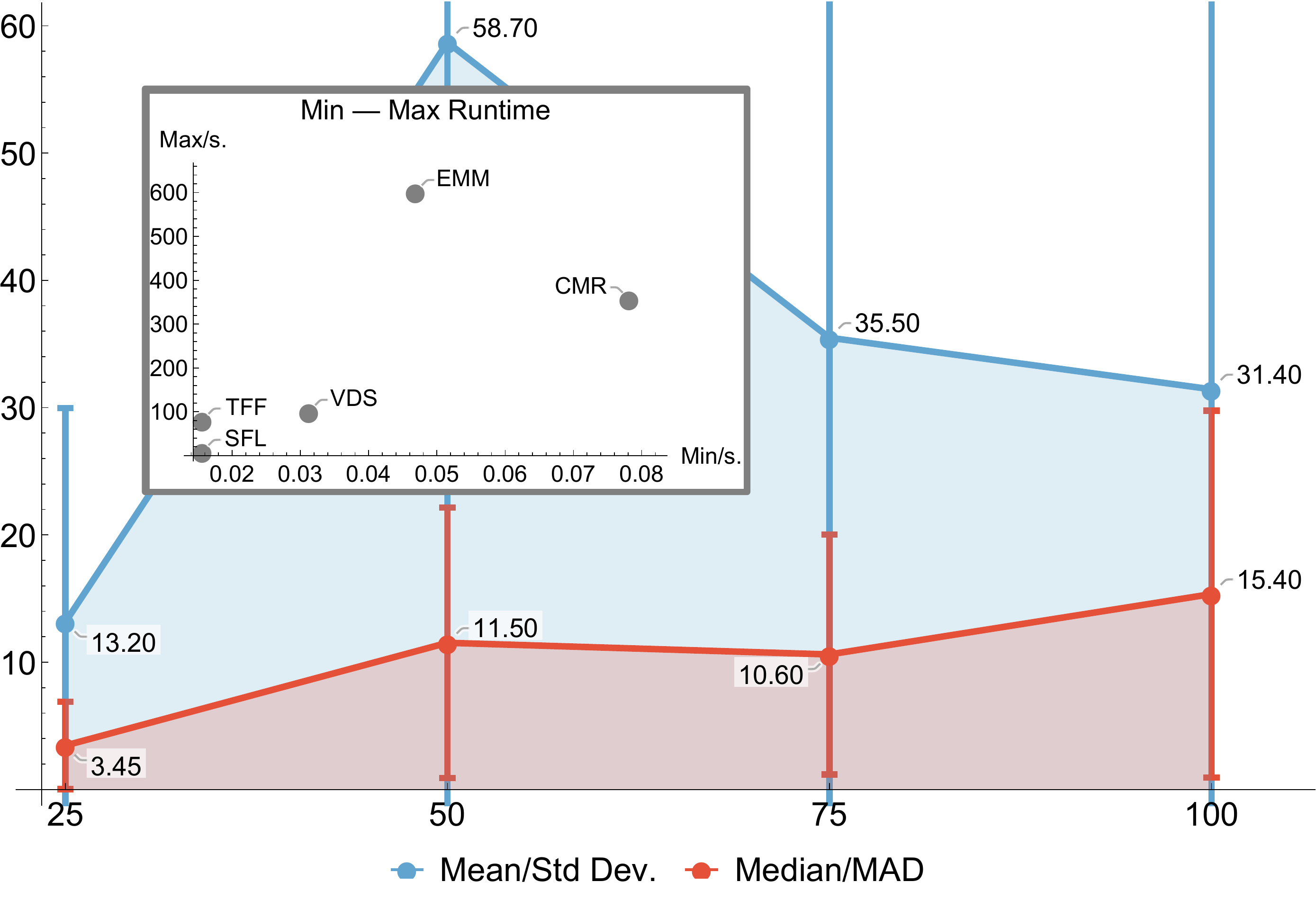} \\
		{\small Legend:} \includegraphics[scale=0.7, valign=m]{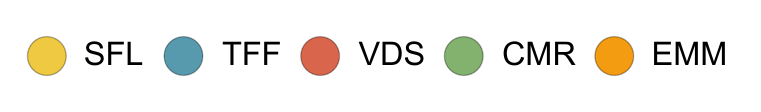} \\	
	\end{tabu}
	\begin{minipage}[t]{\textwidth}
		\footnotesize
		1) The	experiments are applied on $10$ trials for each percent value. In frontside, the mean (in red) and the standard deviation (in gray) are reported for each percent value. In the graphics the minimal (blue) and the maximal (black) distances related to the concerned trials are reported. The diameter of the bubbles indicates the standard deviation ranging from $0.95 \%$ ($75\%$) to $19.00 \%$ ($25\%$) for the binary profile and from $0.63 \%$ ($100\%$) to $15.83 \%$ ($25\%$) for the signature with a different scale to the axis for the sake of clarity. 
		
	\medskip
		2)	The curves show the mean and the median computation time in second of all the networks for the different percentages of profiles and signatures, with the standard deviation and median absolute deviation respectively as error bars. The inserts detail the minimum and maximum duration of each network. The experiments are performed on a quad-core with \textsc{intel} I core 7 \textsc{cpu} and 16~Gb of \textsc{ram} (HP-Zbook 15). 
	\end{minipage}
	\caption{Benchmark results}
	\label{fig:benchmark}
\end{figure}

 Figure~\ref{fig:benchmark}.1 shows the experiments for each trial
 For $0 \%$ of signatures, the formulas are simply randomly selected without computing \taboon.
Figure~\ref{fig:benchmark}.2 reports the computation time 
 and the number of potential models for each reference network. 
 
For all trials and all networks, the optimal score is reached (null vector), meaning that no improvement could be achieved for the solution networks and the failure bound was never reached.
 Therefore the found networks reach the expected signature profiles at equilibria with the same number of stable states as the reference networks. Hence, the solution networks do not differ to reference network for these criteria that are assimilated here the elements validating a model.
The truth-value distance remains low with a mean less than $10\%$ at most ($25\%$) and does not exceed $14.20\%$ (Figure~\ref{fig:benchmark}.1). 
Compared to the random selection of formulas (Signature = $0\%$), \textsc{\taboon} significantly improves the solution since the percent of decrease is respectively $17\%$ for $25\%$ of biomarkers, $40\%$ for $50\%$, and $49\%$ for $75, 100 \%$ with a low initial truth value distance of $5.3\%$ in average.
The increase of the number of biomarkers improves the efficiency of the inference. However, we can remark that the score is the same for $75\%$ and $100\%$.

Finally, the computational time decreases with number of binary profiles (Figure~\ref{fig:benchmark}.2) since this reduces the number of formulas. However no correlations can be drawn between the time and the size of the signature. The median is low compared to the mean stressing the fact that the computation is efficient with the presence of some outlier executions where reaching the optimal score requires many steps. 
 
 \textsc{\taboon} improves the solution compared to a random selection of formulas and the execution time of is reasonable for the tested networks. The obtained truth-value distance indicates that the networks is closed to the dynamics of the reference ones with variation for some entries. This result has to be compared to the number of putative models that corresponds to the product of the number of formulas per nodes (Figure~\ref{fig:benchmark-nb-models}). Among a huge number of putative models \textsc{\taboon} is able to find networks having a near-optimal dynamics. 
Moreover, the major factor of efficiency is related to the amount of data as shown by the evolution of the distance for the binary profiles compared to this of signature (Figure~\ref{fig:benchmark}.1) and the number of putative models (Figure~\ref{fig:benchmark-nb-models}). The increase of the size of the signature also leads to improve the accuracy of the inference until a maximal value that seems to be an asymptote (here $75\%$) suggesting that the choice of the biokmarkers also matters. 
\clearpage
\section{Conclusion}
\label{sec:Conclusion}
Boolean network synthesis is at the core of network based modelling. The synthesis involves a two-stage process: the interaction graph discovery and the dynamical function characterizing the behaviour of the nodes which is a propositional formula for BN. The \taboon method is focused on the second phases. The characterization of a Boolean network occurs in two stages : the formulas inference that are compatible with the Boolean expression profiles for each node and the election of the fittest formula in each node with regards to global properties related to the molecular dynamics of the studied molecular system. This stage is based on a Tabu meta-heuristic for selecting the best formulas using an objective function for formalizing the global biological properties. 

The resulting network gathering all the fittest formulas represents the final outcome of the \taboon work-flow. As the global properties are based on a model validation protocol, the network with an optimal null score also validates the biological observations otherwise the score estimates a distance to the truthful network. Hence the objective function enables the quantification of the accuracy of a network that can reveal useful for improving the model. A network with a null score thus cannot be contradicted with the validation protocol. 
 Using the classical validation criteria as inputs for the Boolean network synthesis constitutes an originality of the method because they are often used to check the validity of a model without being involved for their synthesis.
 
 The experimental assessment shows the efficiency of the method for finding the most truthful network compared to a random selection of formulas. The benchmark also shows the influence of two parameters: the number of Boolean expression profiles and the precision of the global properties for selecting the best candidate network as final outcome. 
 
 It may occur that several networks with an optimal null score can be found. This case can be notably due to the definition of the global properties that insufficiently discriminate the fittest network. Therefore these networks would pave the behavioural space offering alternative models for the analysis. Although only one network should be basically considered as the most truthful, the modelling is thus extended to a pool of networks that cannot be discriminated by the validation process (global properties). Such case addresses as perspective of investigating how to perform relevant modelling with a family of Boolean networks where the remaining uncertainties on the formulas are viewed as unresolved parameters of a network.
 
\clearpage 
\bibliography{bib-Boolean-network-inference}
\bibliographystyle{plain}

%
%

\vfill
\end{document}